# CCKS 2019 知识图谱评测技术报告：实体、关系、事件及问答

韩先培 [1]　王志春 [2]　张江涛 [3,4]　温清华 [4]　李文琪 [5]　汤步洲 [6]　汪琦 [7]
冯知凡 [7]　张扬 [7]　吕雅娟 [7]　王海涛 [8]　陈文亮 [8]　邵浩 [9]　陈玉博 [10,11]　刘康 [10,11]　赵军 [10,11]　王太峰 [12]　张可尊 [12]　王萌 [13,14]　江胤霖 [13]　漆桂林 [13,14]
邹磊 [15]　胡森 [15]　张旻昊 [15]　林殷年 [15]

（1. 中国科学院软件研究所 中文信息处理实验室，北京市 100190；2. 北京师范大学 人工智能学院，北京市 100875；3. 中国人民解放军第三〇五医院，北京市 100017；4. 清华大学 计算机系知识工程实验室，北京市 100084；5. 医渡云（北京）技术有限公司，北京市 100000；6. 哈尔滨工业大学（深圳），广东省深圳市 518055；7. 百度 知识图谱部，北京市 100193；8. 苏州大学 计算机科学与技术学院，江苏省 苏州市 215006；9. 狗尾草智能信息科技有限公司，江苏省苏州市，215002；10. 中国科学院自动化研究所 模式识别国家重点实验室，北京市 100190；11. 中国科学院大学 人工智能学院，北京市 100049；12. 蚂蚁金服（杭州）网络科技有限公司，浙江省杭州市 310000；13. 东南大学，计算机科学与工程学院，江苏省南京市，211189；14. 东南大学 计算机网络和信息集成教育部重点实验室，江苏省南京市，211189；15. 北京大学，北京市 100871）

**摘要：** 知识图谱以概念、实体及其关系的方式建模世界知识，在实际任务中得到了越来越广泛的应用。为了介绍知识图谱的相关技术、评估当前技术的性能水平、了解不同技术的优缺点，揭示未来的发展方向，CCKS 2019 举办了一个包含 6 个任务的评测竞赛，覆盖了实体、关系、事件及问答等多类知识图谱关键技术，吸引了 1600 余支队伍参加。本报告系统整理了参赛队伍使用的技术、资源和策略，可以为评估当前知识图谱技术水平，提供构建知识图谱系统的技术参考，揭示未来的发展方向提供一份有价值的参考。

**关键词：** 知识图谱；评测报告；信息抽取

**中图分类号：** TP391　　　**文献标识码：** A

# Overview of the CCKS 2019 Knowledge Graph Evaluation Track: Entity, Relation, Event and QA

Xianpei Han[1], Zhichun Wang[2], Jiangtao Zhang[3,4], Qinghua Wen[4], Wenqi Li[5], Buzhou Tang[6], Qi Wang[7], Zhifan Feng[7], Yang Zhang[7], Yajuan Lu[7], Haitao Wang[8], Wenliang Chen[8], Hao Shao[9], Yubo Chen[10,11], Kang Liu[10,11], Jun Zhao[10,11], Taifeng Wang[12], Kezun Zhang[12], Meng Wang[13,14], Yinlin Jiang[13], Guilin Qi[13,14], Lei Zou[15], Sen Hu[15], Minhao Zhang[15], Yinnian Lin[15]

(1. Chinese Information Processing Laboratory, Institute of Software, Chinese Academy of Sciences, Beijing 100190, China ; 2. School of Artificial Intelligence, Beijing Normal University, Beijing 100875, China; 3. The 305 Hospital of People's Liberation Army, Beijing 100017, China; 4. Department of Computer Science and Technology, Tsinghua University, Beijing 100084, China; 5. Yidu Cloud, Beijing 100000, China; 6. Harbin Institute of Technology, Shenzhen, Guangdong 518055, China; 7. Baidu, Beijing 100193, China; 8. School of Computer Science & Technology, Soochow University, Suzhou, Jiangsu 215006, China; 9. Gowild , Suzhou, Jiangsu 215002, China; 10. National Laboratory of Pattern Recognition, Institute of Automation, Chinese Academy of Sciences, Beijing 100190, China; 11. School of Artificial Intelligence, University of Chinese Academy of Sciences, Beijing 100049, China; 12. Ant Financial, Hangzhou, Zhejiang 310000, China; 13. School of Computer Science and Engineering, Southeast University, Nanjing, Jiangsu 211189, China; 14. Key Laboratory of Computer Network and Information Integration, Southeast University, Nanjing, Jiangsu 211189, China; 15. Peking University, Beijing 100871, China)

**Abstract :** Knowledge graph models world knowledge as concepts, entities, and the relationships between them,

which has been widely used in many real-world tasks. CCKS 2019 held an evaluation track with 6 tasks and attracted more than 1,600 teams. In this paper we give an overview of the knowledge graph evaluation tract at CCKS 2019. By reviewing the task definition, successful methods, useful resources, good strategies and research challenges associated with each task in CCKS 2019, this paper can provide a helpful reference for developing knowledge graph applications and conducting future knowledge graph researches.

## 0    CCKS 2019 评测总体情况介绍

知识图谱（Knowledge Graph）以概念、实体及其关系的方式建模世界知识，提供了一种更好地描述、组织、管理和理解信息的能力。近年来，由于其有效性，知识图谱被应用于越来越多的应用场景中，解决各种各样的问题。

知识图谱应用具有一个完备的技术体系。在将知识图谱应用于解决实际任务的过程中，往往需要应用到多种知识图谱相关的技术。在上述过程中，了解知识图谱的各种相关技术，评估当前技术的性能水平，了解不同技术的优缺点，参考可行的改进和发展方向，对知识图谱的成功应用至关重要。

CCKS 2019 评测竞赛由中国中文信息学会语言与知识计算专委会举办，其旨在为研究者和应用开发人员提供一个测试技术、算法、及系统的平台，共包含了 6 个评测任务，分别为：任务一、面向中文电子病历的医疗实体识别及属性抽取；任务二、面向中文短文本的实体链指；任务三、人物关系抽取；任务四、面向金融领域的事件主体抽取；任务五、公众公司公告信息抽取；任务六、中文知识图谱问答。

CCKS 2019 评测任务覆盖了实体、关系、事件三类知识图谱的基本知识单元，包括了识别、抽取、链接等关键任务，同时也包含问答这个知识图谱最具代表性的应用。上述任务共吸引了 1666 支队伍报名参加，698 支队伍提交有效结果。

本届评测的参赛队伍从不同策略、不同技术、不同资源的角度对当前知识图谱任务进行攻关。本报告系统整理介绍了参赛队伍（特别是优胜队伍）使用的技术、资源和策略。相信本报告可以为提供构建知识图谱系统的技术参考，评估当前知识图谱技术水平，揭示未来的发展方向提供一份有价值的参考。以下将按任务对评测进行详细介绍。

## 1    任务一：面向中文电子病历的命名实体识别

### 1.1  任务介绍

随着医院信息技术的发展，积累了越来越大规模的电子病历数据，其中以自由文本形式的非结构化数据是电子病历最主要的部分，如出院小结、病程记录等，其中蕴含着大量详尽而有价值的医疗知识及健康信息[1]。从这些海量的电子病历文本中识别出与医疗相关的实体名称，并将它们归类到预定义类别，如疾病、治疗、症状、药品等，是电子病历数据挖掘与信息抽取的关键步骤，这一任务通常称之为面向电子病历的命名实体识别[2]。它不仅是 NLP 相关任务如信息检索、信息抽取以及问答系统等的重要基础工作[3]，同时对诸多实际临床应用如合并症分析、综合症监测、不良药物事件检测以及药物相互作用的分析等具有巨大促进作用[4]。

国际上先后出现了一些围绕电子病历命名实体识别的公开评测及标注数据集，主要有 i2b2[5]，ShARe CLEF eHealth[6]以及 SemEval[7]，但这些评测大多面向英文电子病历。为了更好的促进中文电子病历相关研究的发展，填补国内面向中文电子病历命名实体识别评测竞赛及标注数据集的空白，清华大学知识工程实验室联合医渡云北京科技有限公司、哈尔滨工业大学（深圳）组织了这次评测任务。

**任务定义** 本任务是全国知识图谱与语义计算大会（CCKS）围绕中文电子病历语义化开展的系列评测的一个延续，包括两个子任务：

（1）**医疗命名实体识别**：CCKS 2017[8]，2018[1]评测任务的延续，对 2018 年度数据集做了标注修订。形式化定义如下：

**输入**：

1. 电子病历的自然语言文本集合：
$\mathcal{D} = \{d_1, \cdots d_N\}$，$d_i = \langle w_{i1}, \cdots w_{in} \rangle$

2. 预定义类别：$C = \{c_1, \cdots c_m\}$

**输出**：

实体提及和所属类别对的集合：
$$\{\langle m_1, c_{m_1}\rangle, \langle m_2, c_{m_2}\rangle, \cdots \langle m_p, c_{m_p}\rangle\}$$

其中 $m_i = \langle d_i, b_i, e_i\rangle$ 是出现在文档 $d_i$ 中的医疗实体提及（mention），$b_i$ 和 $e_i$ 分别表示 $m_i$ 在 $d_i$ 中的起止位置，$c_{m_i} \in C$ 表示所属的预定义类别。要求实体提及之间不重叠，即 $e_i < b_{i+1}$。

本任务预定义类别包括：疾病和诊断、检查、检验、手术、药物、解剖部位六大类。

（2）**医疗实体及属性抽取（跨院迁移）**：在医疗实体识别的基础上，对预定义实体属性进行抽取。该子任务为迁移学习任务，即在只提供目标场景少量标注数据的情况下，通过其他场景的标注数据及非标注数据进行目标场景的识别任务。形式化定义如下：

输入：

1. 电子病历的自然语言文本集合：$\mathcal{D} = \{d_1, d_2 \cdots d_N\}$, $d_i = \langle w_{i1}, w_{i2} \cdots w_{in}\rangle$

2. 预定义类别：$P = \{p_1, p_2 \cdots p_m\}$

输出：

预定义类别的答案实体集合：
$$\{\langle d_i, \langle p_j, \langle s_1, s_2, \cdots s_k\rangle\rangle\rangle\}, 1 \le i \le N, 1 \le j \le m$$

其中 $s_k$ 为 $d_i$ 中出现的属于 $p_j$ 答案实体（字符串，无需位置信息），每个类别可包含 0 或多个实体。

**组织安排** 本任务由清华大学知识工程实验室统筹组织，创建了专门的评测交流讨论组：CCKS2019-clinic@googlegroups.com。时间安排如下：

- 评测任务发布： 4月1日
- 报名时间：4月1日—7月10日
- 训练数据发布（第一批）：5月6日
- 训练数据发布（第二批）：6月2日
- 测试样例数据发布：7月20日
- 测试数据发布及测试结果提交：7月30日
- 评测论文提交：8月15日

评测结果在 biendata 网站[1]上进行在线提交，采用实时打榜的方式进行。最终由任务组织者综合考虑网站得分、结果可复现性（要求参赛队提交代码及文档）、方法模型及外部资源（词典、非本评测提供的标注数据等）等因素给出最终成绩排名。

**参赛情况** 本次评测共吸引了来自院校科研机构、工业界及医疗机构的 200 多支队伍参赛。参赛情况如表 1 所示。

表 1 参赛情况

| 任务 | 报名数 | 结果提交 | 代码提交 | 论文 |
|---|---|---|---|---|
| 1 | 226 | 44 | 10 | 7 |
| 2 | 152 | 6 | 6 | 4 |

## 1.2 数据集及评估方法

**数据集描述** 两个子任务的数据集均由医渡云（北京）技术有限公司提供，并组织专业的医学团队进行人工标注，仅限 CCKS 竞赛评测用[2]。

数据集的统计信息如表 2 和表 3 所示。

**评估方法** 子任务 1 采用传统的精确率（Precision）、召回率（Recall）以及 F1-Measure 作为评测指标。参赛系统的输出结果集合记为 $S = \{s_1, s_2 \ldots s_m\}$，人工标注的结果（Gold Standard）集合记为 $G = \{g_1, g_2 \ldots g_n\}$。集合元素为一个实体提及，表示为四元组 $\langle d, pos_b, pos_e, c\rangle$，$d$ 表示文档，$pos_b$ 和 $pos_e$ 分别对应实体提及在文档 $d$ 中的起止下标，$c$ 表示实体提及所属预定义类别。遵循国际惯例，分别从两个层面进行评价。

**严格指标** 我们定义 $s_i \in S$ 与 $g_j \in G$ 严格等价，当且仅当：

1. $s_i.d = g_j.d$
2. $s_i.pos_b = g_j.pos_b$
3. $s_i.pos_e = g_j.pos_e$
4. $s_i.c = g_j.c$

基于以上等价关系，我们定义集合 $S$ 与 $G$ 的严格交集为 $\cap_s$。由此得到严格评测指标：

$$P_s = \frac{|S \cap_s G|}{|S|}, \quad R_s = \frac{|S \cap_s G|}{|G|}, \quad F_{1s} = \frac{2PR}{P+R}$$

**松弛指标** 我们定义 $s_i \in S$ 与 $g_j \in G$ 松弛等价，当且仅当：

1. $s_i.d = g_j.d$
2. $\max(s_i.pos_b, g_j.pos_b) \le \min(s_i.pos_e, g_j.pos_e)$ [3]
3. $s_i.c = g_j.c$

基于以上等价关系，我们定义集合 $S$ 与 $G$ 的松弛交集为 $\cap_r$。由此得到松弛评测指标：

$$P_r = \frac{|S \cap_r G|}{|S|}, \quad R_r = \frac{|S \cap_r G|}{|G|}, \quad F_{1r} = \frac{2PR}{P+R}$$

子任务 2 由于每个文本的一个目标字段（类别）可能出现多个实体，评测指标使用实体而非类别来计算准召率，最终使用实体的 F1 值为评测指标。

---

表 2 子任务 1 数据集描述

| 数据集 | 文档数 | 疾病和诊断 | 检查 | 检验 | 手术 | 药物 | 解剖部位 | 总数 |
|--------|--------|-----------|------|------|------|------|---------|------|
| 训练集 | 1000 | 2116 | 222 | 318 | 765 | 456 | 1486 | 5363 |
| 测试集 | 379 | 682 | 91 | 193 | 140 | 263 | 447 | 1816 |

表 3 子任务 2 数据集描述

| 数据集 | 文档数 | 目标场景 | 非目标场景 | 肿瘤部位 | 肿瘤大小 | 转移部位 | 实体总数 |
|--------|--------|---------|-----------|---------|---------|---------|---------|
| 训练集 | 1000 | 100 | 900 | 180 | 454 | 602 | 1236 |
| 测试集 | 400 | 400 | – | 112 | 95 | 165 | 372 |
| 非标注 | 1000 | x | 1000–x | – | – | – | – |

## 1.3 典型模型与系统

通过对本次评测所提交的 11 篇论文进行分析，将所采用的典型方法、模型总结如下。

**序列标注模型：** 无论子任务 1 还是 2，都是对预定义类别（属性）的实体进行识别和归类，因此，所有参赛队都将任务视为序列标注问题，采用的主体模型都是 CRF、BiLSTM 等传统的序列标注模型。

**混合模型：** 融合方法是主流，大部分参赛队以多种方式融合了多个模型，如 CRF,BiLSTM+CRF，以及 BiLSTM+CNN+CRF 等，相比于单一模型获得了很大的性能提升。

**预训练语言模型：** 基于 Bert、ELMo 等语言模型在自然语言处理任务中的卓越表现，大多数参赛队在文本预训练中引入 Bert 等语言模型,评测结果证明这些语言模型的引入能够有效提升医疗命名实体识别的性能。

**特征工程及规则定义：** 尽管表示学习方法在通用领域中取得了很大的成功，但是针对临床医疗领域的具体任务，特征工程依然不可或缺，所有的参赛队都进行了大量的特征定义，包括词性标注、拼音特性、词根、偏旁部首以及词典特征等，并搭配使用人工定义规则进行预处理和后处理来提升性能，间接反映出单纯采用嵌入表示、无特征工程的方法在实际的临床医疗文本中并不适用，一个重要的原因在于无法获得高质量大规模的公开电子病历语料进行表示学习训练。

## 1.4 评测性能及结果分析

最终评测结果如表 4、表 5 所示（排名前 6）。

表 4 子任务 1 评测结果

| 排名 | 参赛队名 | 单位 | 得分 |
|------|---------|------|------|
| 1 | Alihealth | 阿里健康科技有限公司 | 0.85620 |
| 2 | THU_MSIIP | 清华大学科大讯飞联合实验室 | 0.85592 |
| 3 | DUTIR | 大连理工大学 | 0.85162 |
| 4 | jfhealthcare | 江西中科九峰智慧医疗有限公司 | 0.84846 |
| 5 | suda-hlt | 苏州大学 | 0.84121 |
| 6 | ZJUCST | 浙江大学 | 0.83795 |

表 5 子任务 2 评测结果

| 排名 | 参赛队名 | 单位 | 得分 |
|------|---------|------|------|
| 1 | NUDT-YH | 国防科技大学 | 0.76350 |
| 2 | THU_MSIIP | 清华大学科大讯飞联合实验室 | 0.76165 |
| 3 | DUTIR | 大连理工大学 | 0.70490 |
| 4 | zu_nlp | 中原工学院 | 0.63331 |
| 5 | SCNU_TAMlab | 华南师范大学 | 0.59906 |
| 6 | 四道口队 | 北京交通大学 | 0.59253 |

通过对参赛队所提交的代码，文档及评测论文的对比审核，评测结果分析如下。

1. 子任务 1 排名前 10 的参赛队，得分非常接近，如排名第一、第二的参赛队 F1 分差只有 0.00028，他们所采用的模型主体部分都是 BiLSTM+CRF，方法差异性较小。由于单纯依

靠模型和算法很难获得显著的性能提升，参赛队大多依靠模型之外的规则定义、特征工程等人工干预提升性能。排名第一的参赛队更是通过对训练标注数据集进行人工标注修正后，合并两个数据集进行混合训练，取得了最好的成绩。

2. 子任务 2 的整体性能要远低于子任务 1，一个根本的原因在于它是一个迁移学习任务，给出的训练数据集与最终的测试数据集来自于完全不同的实际场景，而所有的参赛队都没有考虑场景迁移问题，且对评测中给出的1000 条非标注数据也没有进行充分利用，导致性能较差。

3. 与往年评测[8][1]相比，一个显著的改进是随着 2018 年以来语言模型（language model）在自然语言处理任务中的卓越表现，所有的参赛队在预训练中都引入了 Bert, ELmo 等模型，结果表明这些语言模型能有效提升识别性能。

4. 由于医疗实体名称的特殊性，词典的构造至关重要，在所提交的评测论文中，都借助大量外部词典资源，如 ICD-10 疾病编码，国际标准手术编码，药品名称大全，解剖学词库等，并从"寻医问药"、"好大夫"等网站上收集大量的专业术语，构建多个类型的术语词典进行匹配。

5. 由于医疗文本的特殊性，所有的参赛队都摒弃了分词预处理环节，借助评测提供的训练集和测试集以及外部语料库，如百度百科等，直接训练字嵌入向量（character embedding），而将分词和命名实体识别任务进行联合建模（joint model）的方法还没有出现。

本任务共收录了 6 篇评测论文（两个子任务各 3 篇），有兴趣的读者可参见 CCKS 2019 评测论文集[^4]。

### 1.5 任务总结

本评测是围绕中文电子病历语义化开展的一系列评测的起点，为后续即将推出的医疗实体结构化、医疗实体标准化以及医疗实体关系抽取、医疗知识图谱构建等任务打下坚实基础。

从参赛结果看出，相比于面向英文电子病历的国际评测，如 i2b2 以及 ShARe CLEF eHealth，参赛队大多都取得了良好的成绩。一个主要的原因是中英文数据集上实体分布不同，中文数据集中实体重现率更高。但在方法和模型的创新上稍

显不足，我们认为基于中文电子病历命名实体识别工作仍值得进一步研究。

我们希望通过组织这次评测，推动该领域的研究进展和技术实用化，活跃学术氛围，促进各个单位和团体的交流，希望在以后的评测中能够做得更好。

## 2 任务二：面向中文短文本的实体链指

### 2.1 任务介绍

面向中文短文本的实体链指（Entity Linking），是自然语言处理领域的基础任务之一，即对于给定的一个中文短文本（如搜索 Query、微博文本、用户对话内容、文章标题等）识别出其中的实体，并与给定知识库中的对应实体进行关联的过程。实体链指包括实体识别(NER)和实体消歧(NED)两个子任务。

实体链指可用于搜索、推荐、广告、对话等多种场景下的知识解析任务，同时可以在很多任务中发挥辅助作用，如应用于文本理解、意图理解、舆情分析、对话 NLU 等任务中，用于提升任务精度。

传统的实体链指任务主要是针对长文档，长文档拥有较长的上下文信息能辅助实体的歧义消解并完成链指。相比之下，针对中文短文本的实体链指存在很大的挑战，主要原因如下：

（1）中文短文本语料口语化严重，导致实体歧义消解困难；

（2）短文本上下文语境不丰富，须对上下文语境进行精准理解；

（3）存在命名多样性、实体多歧义的问题，相比于英文，中文由于语言自身特点，在短文本的实体链指问题上更有挑战。

本次实体链指评测共吸引来自 4 个国家的 345 支队伍、862 名参赛者报名参加：分别来自165 个院校和 67 家企业，最终共有 154 支队伍提交结果，整个比赛周期内累计提交评测结果 1203次。

### 2.2 数据集及评估方法

（1）知识库：

本任务的参考知识库来自百度百科知识库的约 39 万个实体。知识库中的每个实体都包含一个实体 id(subject_id)，字符串名称(subject)，上位概念（type）及与此实体相关的一系列属性和属性值<predicate, object>形式。知识库中每行

---

[^4]: https://conference.bj.bce-bos.com/ccks2019/eval/webpage/index.html

代表知识库的一条记录，每条记录的格式为一个 Json 格式。`subject_id` 的值为一个正整数，说明 `subject` 总是对应知识库中的一个实体。

知识库数据分布如表 6 所示。

表 6 知识库数据分布

| 类型 | 数量 |
| --- | --- |
| 实体数量 | 398,028 |
| SPO 数量 | 3,564,565 |
| 实体描述数据 | 361,778 |
| 实体平均属性数量 | 9 |
| 实体描述数据平均长度 | 103 |

（2）数据集：

标注数据集由训练集、验证集、评测集组成。其中训练集中包括 9 万条短文本标注数据，验证集包括 1 万条短文本数据，评测集包含 3 万条短文本数据，所有数据通过人工众包方式标注生成。

标注数据集主要来自于真实的互联网网页标题数据，是用户检索 Query 对应的有展现及点击的网页，短文本平均长度为 21.73 中文字符，覆盖了不同领域(如人物、电影、电视、小说、软件、组织机构、事件等垂类)的实体以及通用概念。

（3）评估方法：

我们以 F1 分值作为评价指标，对于给定的中文短文本，实体链指策略输出的结果中包含给定中文短文本中出现的所有命名实体的链指结果。我们通过将输出结果与人工标准集合进行比较来计算准确率(Precision)，召回率(Recall)和 F1 分值(F1 score)。

具体计算过程如下所示：

给定短文本输入（用 Text 表示，其属于 golden 标注集），此 Text 中有 N 个实体 mention：$M_n = \{m_1, m_2, m_3 \ldots m_n\}$，每个实体 mention 链接到知识库的实体 id 为：$E_n = \{e_1, e_2, e_3 \ldots e_n\}$，实体标注系统输出标注结果如下：$E'_n = \{e'_1, e'_2, e'_3 \ldots e'_n\}$，则实体标注的准确率定义如下：

$$P = \frac{\sum_{n \in N} |E_n \cap E'_n|}{\sum_{n \in N} |E'_n|}$$

实体标注的召回率定义如下：

$$R = \frac{\sum_{n \in N} |E_n \cap E'_n|}{\sum_{n \in N} |E_n|}$$

实体标注的 F1 值定义如下：

$$F1 = \frac{2 * P * R}{P + R}$$

## 2.3 典型模型与系统

实体链指任务主要可拆解为实体识别和实体消歧两个关键子任务，实体链指的传统做法是采用管道的方式，先进行实体识别，然后将实体识别的结果作为实体消歧的输入；另外一种做法是采用端到端(end2end)的方式训练一个端到端模型，输入是一段文本，输出即是实体链指的结果，本次评测绝大多数选手采用的是第一种方法。

下面对本次评测使用频率较高且相对典型的模型与方法进行介绍：

（1）实体识别：

方法一：基于 CRF[9] 进行序列标注的实体识别方法

如图 1 所示，(a)基于 embedding+LSTM[10] + CRF 的实体识别方法；(b) 在 ERNIE[11]/BERT[12] 等语言模型进行微调(fine-tune) + CRF 的实体识别方法。

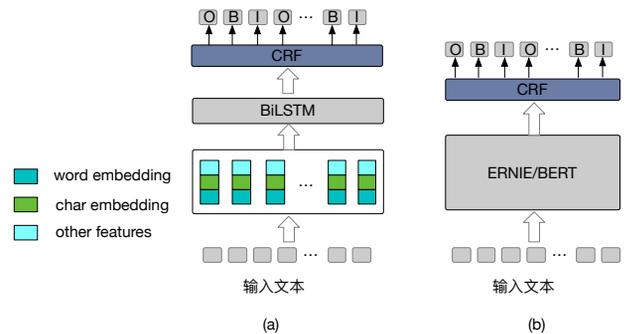

图 1 基于序列标注的实体识别

方法二：基于双指针标注的实体识别方法

如图 2 所示，主要采用双指针标注的方式，通过建模并标注实体的开始位置和终止位置。

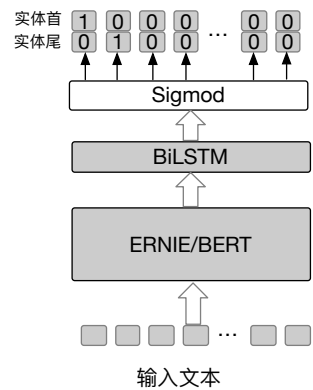

图 2 基于双指针标注的实体识别

（2）实体消歧：

方法一：基于多分类的实体消歧方法：

如图 3 所示，将输入短文本和待消歧实体的描述文本分别输入到 ERNIE/BERT，将输入短文本 CLS（Special Classification Embedding，用于分类的向量，会聚集所有的分类信息，一般是整体序列的向量表示）位置输出向量和实体文本 CLS 位置的输出向量连接到一起得到实体的向量表示，经过 Dropout 层和全连接层，最后进行

Softmax 多分类。

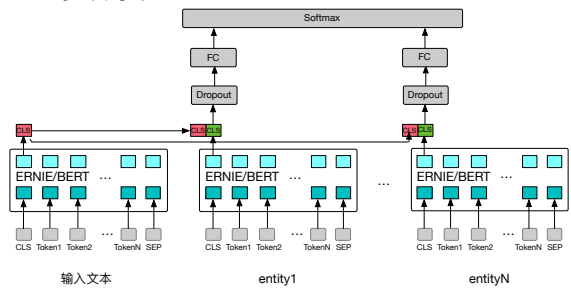

图 3 基于多分类的实体消歧模型

方法二：基于二分类的实体消歧方法：

如图 4 所示，将输入短文本和待消歧实体的描述文本拼接，输入到 BERT，将 CLS 位置输出向量以及候选实体对应开始和结束位置对应的特征向量连接到一起，经过全连接层，最后 Sigmoid 激活得到候选实体的概率得分。

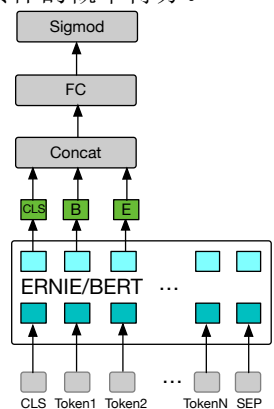

图 4 基于二分类的实体消歧模型

方法三：基于排序的实体消歧方法：

如图 5 所示，使用语义匹配的思路，计算待消歧文本（输入文本）和知识库中候选实体间的相似度，通过神经网络对它们分别建模并得到各自的向量表示，然后通过相似度度量方法进行匹配度打分，选择得分最高的候选实体输出。

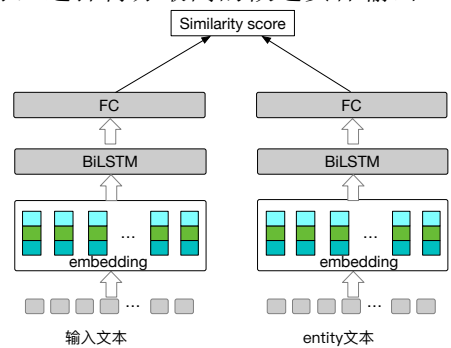

图 5 基于排序的实体消歧模型

## 2.4 评测性能及结果分析

如表 7 所示，表 7 展示了本次评测 Top10 团队的相关信息和得分情况，其中 Top7 团队的 F1 值都达到了 0.79 以上，Top5 团队均在模型中引入 BERT、ERNIE 等预训练语言模型并取得了良好的效果，同时 Top10 队伍中亦有多个团队通过各种形式引入知识来提升整体模型的效果，如基于知识图谱的实体嵌入（entity embedding）和实体 mention 嵌入（mention embedding）等，知识增强对模型效果的提示发挥着非常关键的作用。

同时，我们也发现，模型融合仍是提升算法效果的重要手段，Top5 团队都采用了模型融合来提升整体效果。

表 7 Top10 团队信息及得分

| 排名 | 队伍名 | 参赛单位 | F1 值 |
|---|---|---|---|
| 1 | Free | 东北大学 | 0.80143 |
| 2 | Team KG | 联想研究院人工智能实验室 | 0.79965 |
| 3 | 观 | 电子科技大学 | 0.79654 |
| 4 | zhuflower | 上汽集团人工智能实验室 | 0.79326 |
| 5 | 烟雾弹大师法棍诺 | 同济大学 | 0.79266 |
| 6 | Jun | 国防科技大学 | 0.79048 |
| 7 | BOE_IOT_AIBD | 京东方科技集团股份有限公司 | 0.79023 |
| 8 | hellojet | 浙江大学&合肥工业大学&电子科技大学 | 0.78586 |
| 9 | chill | 上海财经大学 | 0.78483 |
| 10 | ★= base-line =★ | 万达信息股份有限公司&个人 | 0.78450 |
| | JayPath | 中国科学技术大学 | 0.78450 |

## 2.5 任务总结

通过本次评测，参赛选手对实体链指技术做了很多有益探索。知识增强在实体链指中发挥着巨大作用，如基于知识图谱的知识嵌入技术、基于知识图谱的实体动态概念预测技术等，以上技术充分利用了知识来辅助实体消歧任务。

模型融合可以体现集体智慧，也是在比赛中被大家普遍采用的方式，我们发现在评测中获得名次较高的队伍都不同程度的使用了模型融合方法来提升效果。

目前大部分的思路仍然将实体链指任务拆分成实体识别、实体消歧的两个独立的步骤，每个步骤作为一个独立的技术任务，最后进行串联，而近年来提出的利用深度神经网络的端到端解决方案，可以利用实体识别和实体消歧间的信息交

互进行整体建模，可能是进一步提升效果的新思路。

随着预训练语言模型（BERT、ERNIE 等）的快速发展，一些基于预训练语言模型的零样本学习的实体链指方案[13]也被提出，而其本身也是充分利用知识来解决样本问题和跨领域的泛化问题。

分析与研究表明，实体链指任务面临着开放域、行业场景、少样本学习带来的新挑战，结合相应的问题场景，探索更多更好的知识增强技术，运用神经网络的强大学习建模能力进一步提升实体链指技术效果可能是未来研究的主要方向。

# 3 任务三：人物关系抽取

## 3.1 任务介绍

作为 CCKS2019 的一项评测任务，人物关系抽取的目标是，从给定的句子中识别出指定人物实体对的关系。根据评价方式的不同，本次评测任务又分为两个子任务：Sent-Track 和 Bag-Track。在 Sent-Track 中，关系预测是在句子级别上进行的。而在 Bag-Track 中，关系预测是在包级别上进行的，所有包含相同实体的句子组成一个包。在本次评测任务中，一共有 358 支队伍报名参加，其中 147 支队伍提交了有效成绩。

人物关系抽取是一个非常有潜力的研究方向。一方面，关系抽取作为信息抽取的一个重要子任务，是智能问答、信息检索等许多智能应用的重要基础，和知识图谱的构建有着密切的联系；另一方面，人作为社会的重要组成单位，研究人与人之间的关系是十分必要和有价值的。[14]

目前，有监督的关系抽取方法[15,16,17,18]被广泛使用，并且表现出了非常好的性能。但是，有监督的方法往往需要大量的标注数据用于训练，人工标注训练数据是十分耗时费力的。为解决这个问题，Mintz 等人[19]提出了远程监督的概念。他们认为，如果两个实体$<e_h, e_t>$在知识库中具有关系$r$，那么所有包含这两个实体的句子都会在某种程度上表达他们的这种关系。通过远程监督的方法，我们可以通过对齐知识库和文本，自动地构建大量标注数据。

然而，通过远程监督方法生成的数据不可避免地存在错误标注的问题，特别是测试数据的错误会导致在比较模型性能时出现错误评估的问题。为了解决这个问题，我们可以让标注人员对测试数据进行标注。本次评测任务中，我们使用一个专门用于人物关系抽取的中文数据集(Inter-Personal Relationship Extraction，IPRE)[20]。在这个数据集中，验证集和测试集是经过人工标注的，训练集是远程监督自动生成的。

## 3.2 数据集及评估方法

本次评测任务的文本数据主要来自于互联网网页文本，通过远程监督自动标注和人工标注相结合的方法生成标注数据。由于目前没有像 Freebase 这样的中文知识库提供足够多的人物关系三元组用于文本对齐，我们需要从维基风格的网页上抽取三元组。通过实体的类别、名字长度等多种方式对网页中的所有实体进行过滤，最后我们构建了一个包含 942,344 个人物实体的人名表。我们根据人名表，进一步挑选网页信息框（infobox）中三元组，这些三元组的实体都在人名表中出现。我们对三元组中出现的关系表达进行合并和去噪，并定义了一组包含 34 种人物关系的集合。最后，我们将这些三元组对齐文本，得到了超过 41,000 个句子和 4,214 个包。

表 8 IPRE 中数据的划分与标注

| 数据集 | 比例 | 包 | 远程监督 | 人工标注 |
|---|---|---|---|---|
| 训练集 | 70% | 2948 | √ | × |
| 验证集 | 10% | 416 | √ | √ |
| 测试集 | 30% | 850 | √ | √ |

如表 8 所示，在得到远程监督生成的数据后，我们将其按 7:1:2 的比例切分成训练集、验证集和测试集。其中，验证集和测试集是经过人工标注的。为了模拟真实的关系抽取场景，我们对各个部分的数据集引入了大量自动构建的 NA 数据。NA 关系是一种特殊的关系，表示实体对不存在指定的人物关系表中的关系。因此，验证集和训练集中的 NA 数据除了少部分来源于人工标注，大部分是来源于后期数据构建时引入的自动构建数据。

在 Sent-Track 和 Bag-Track 两个子任务中，我们都使用$F_1$值作为评价指标。在计算$F_1$值时，我们不考虑 NA。具体来说，如果一个句子或包被预测有多个关系，那么它是否预测有 NA 关系不会影响$F_1$值的计算。记$N_{sys}$是系统预测结果中所有句子或包的非 NA 关系数量，$N_{std}$是标准答案中非 NA 关系的数量，$N_r$是预测正确的关系数量，则 Sent-Track 和 Bag-Track 的评价指标可定义为：

$$P = \frac{N_r}{N_{sys}}, R = \frac{N_r}{N_{std}}, F_1 = \frac{PR}{2PR}$$

在本次评测中，Sent-Track 和 Bag-Track 都有一个公开成绩排行榜（A 榜）和非公开成绩排行榜（B 榜）。所有的测试数据一开始就全部发布，

测试数据按 1:1 的比例切分成了两部分。对于参赛队伍提交的每份预测结果，我们在其中固定的 50%测试集数据和全部测试集数据上计算 F1 值，分别记为 A 分和 B 分。在比赛期间，我们只根据 A 分显示 A 榜，而对应的 B 分进行排名得到 B 榜。B 榜在比赛结束后公布，作为各个参赛队伍的最终成绩。

表 9  Sent-Track 和 Bag-Track 前十名成绩

| Sent-Track | | | | Bag-Track | | | |
|---|---|---|---|---|---|---|---|
| 排名 | 队伍 | A 分 | B 分 | 排名 | 队伍 | A 分 | B 分 |
| 1 | 格物致知 | 0.54076 | 0.54279 | 1 | LEKG | 0.59925 | 0.63030 |
| 2 | LEKG | 0.47300 | 0.48427 | 2 | 格物致知 | 0.60773 | 0.62162 |
| 3 | NEU_DM1 | 0.44912 | 0.46200 | 3 | NEU_DM1 | 0.55894 | 0.57459 |
| 4 | LMN | 0.41171 | 0.41096 | 4 | Ac | 0.51899 | 0.53196 |
| 5 | RE 小分队 | 0.42841 | 0.41003 | 5 | idke_NEU | 0.49754 | 0.52374 |
| 6 | runit | 0.40297 | 0.39523 | 6 | Jun | 0.50785 | 0.52346 |
| 7 | guanchong | 0.38566 | 0.38657 | 7 | OneOf | 0.48895 | 0.50165 |
| 8 | Jun | 0.40322 | 0.38044 | 8 | jack | 0.49351 | 0.49038 |
| 9 | 机器没有命运 | 0.37741 | 0.35767 | 9 | guanchong | 0.45783 | 0.47665 |
| 10 | uw1 | 0.35834 | 0.34885 | 10 | 华凌 NLP | 0.44852 | 0.47612 |

## 3.3 典型模型与系统

自从 BERT[12]发布后，它很快超过 ELMo[21]，成为目前 NLP 领域最火的一个语言模型，在许多 NLP 任务中表现出了非常好的性能。和传统的训练词向量的方法不同，例如 Skip-Gram[22] 和 GloVe[23]，BERT 可以提供更丰富、语义更强的上下文表示。在这次评测任务中，BERT 输出后简单地添加一个全连接层进行分类或者用 BERT 的代替词向量作为输入的两种主流方法都取得了非常好的成绩。考虑到 BERT 可能会过拟合实体的名称，LEKG 队在使用 BERT 时用两个固定的名字"刘伟明"、"李静"替代了句子中的给定的人物实体对。

由于 Sent-Track 和 Bag-Track 使用同样的数据集，并且包级别的关系抽取方法有更好的抗噪性，许多队伍将 Bag-track 的预测结果转换成了 Sent-track 的预测结果。任务转换的关键在于检查包中的句子是否表达了包的这种关系。训练一个二分类器或者使用特征模板匹配的方式都是非常好的选择。在本次评测任务中，NEU_DM1 队和格物致知队都使用了这种转换技巧，他们的实验结果表明这个方法是简单有效的。

特征工程是提高关系抽取系统性能的一种简单有效的方法，需要对原数据进行仔细的分析。就人物关系而言，由于数据本身的特点，有一些特征如性别、姓氏，可以有效地利用进行人物关系的识别和预测。目前一些基于中文人名进行人

物性别预测的开源工具[5]已经取得了不错的效果，姓氏也可以用于血亲关系尤其是父子关系的预测上来。此外，也可以通过一些句子中的关键词对关系的预测结果进行纠正，如"夫"、"妻"等。

考虑到数据分布的不平衡性，一些队伍如 NEU_DM1 队删除了近一半的关系，以减少这部分数据对整体预测结果的影响，他们的实验结果也证明这种思路是有效的。此外，为了扩充训练语料，RE 小分队使用翻译工具[6]对训练数据进行翻译，再将翻译的结果再翻译回中文从而扩充训练数据。针对数据不平衡的问题，格物致知队则使用下采样和上采样相结合的方法数据进行预处理，从而在一定程度上缓解了这个问题。

## 3.4 评测性能及结果分析

在表 9 中，我们给出了 Sent-Track 和 Bag-Track 前十名的成绩。值得注意的是，因为我们的排名策略，最后排名有一些变化。

正如前面提到的那样，预测结果中是否有 NA 关系不会影响 $F_1$ 值的计算，并且评测数据中各关系的分类也是不平衡的。这就导致，如果一个关系抽取系统对其中几个大类的预测有非常好的性能时，在整体数据上的性能也同样会很好。在测试集数据中，"现夫"、"现妻"、"生父"、"生母"、"儿子"、"恋人"、"老师"这七种关系占据了大多数。在 Sent-Track 中，2300 个标准答案中有 2000 个属于这七类；在 Bag-Track 中，740 个标准答案

---

[5] https://pypi.org/project/ngender      [6] https://fanyi.baidu.com/

中有 560 个属于这七类。

我们分别分析了 Sent-Track 和 Bag-Track 前三名的结果，他们都在一定程度上放弃了对一些小类别关系的预测，而把重点放在一些大类关系的预测上。我们通过 $F_1$ 值比较了每个系统在各关系上性能，所有的系统都在这七类关系中的大多数关系上取得了很好的效果。大致可以看出，这几类的关系的预测结果和整体的预测结果是正相

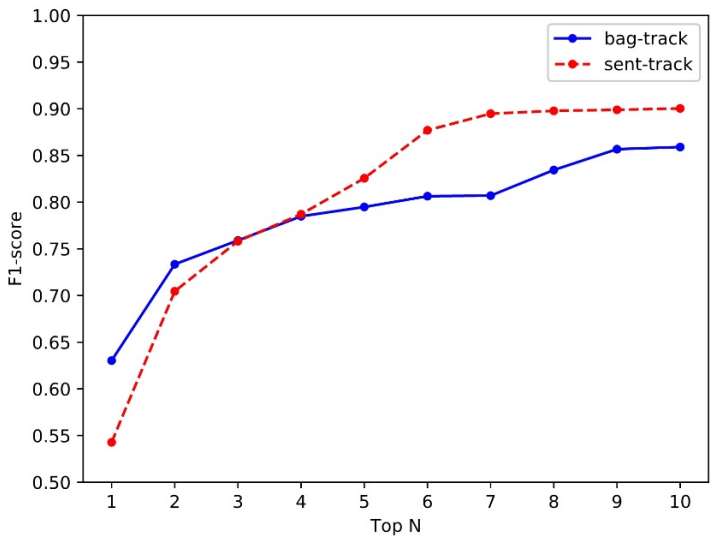

图 6  Top-N 系统预测结果融合的上界

关的。值得注意的是，这几个系统都在"恋人"关系的预测上表现得非常差。我们随机的采样了测试集中一些能体现"恋人"关系的句子，尽管这些句子中大都有"相恋"、"男友"、"女友"这些关键词，但更倾向于把它们预测为 NA。另外，前三名的系统更喜欢把 NA 预测为"生父"。人工检查表明许多只能表达"父母-子女"关系的句子被错误地预测成了"生父"。当然，测试集中引入的远程监督生成的 NA 数据中也产生了一定的影响。

为了给以后的研究提供一个参考上界，我们融合了前十名队伍系统的预测结果。对于每个包或句子，只要其中一支队伍给出了正确答案，我们就认为是正确的。如图 6 所示，我们可以看出，随着融合的结果增加，F1 值的提升明显。令人惊讶的是 Sent-Track 的上界最后超过了 Bag-Track 的上界。这可能是因为一个包往往有多个关系，相比于只有一个关系的句子预测，包预测的难度更大一些。

### 3.5  任务总结

CCKS2019 人物关系抽取评测由 Sent-Track 和 Bag-Track 两个子任务组成。任务目标是识别所给句子中两个人物实体的关系。今年，来自不同单位组织的 358 支队伍参加了比赛。其中，Sent-Track 和 Bag-Track 两个子任务中最好的系统的

F1 值分别达到了 54.3% 和 63.0%。本次评测任务所使用的数据集及其标准答案可以在 https://github.com/SUDA-HLT/IPRE 下载。

## 4  任务四：面向金融领域的事件主体抽取

### 4.1  任务介绍

**任务背景**：事件抽取是信息抽取的重要任务，也是知识图谱自动构建、文本语义理解、自动问答、舆情监控等多种自然语言处理任务的基础。事件知识、特别是事件的类型和主体知识，在金融领域中具有重要作用，能为风险预警、智能投顾等应用提供重要决策参考。在风险预警方面，风险事件的主体识别可以辅助事件热度、事件演化和事件影响的监控，从而进行相应的决策和响应。在智能投顾方面，事件是投资分析和资产管理的重要决策因子，事件主体的识别能辅助发现潜在客户，为智能推荐和投顾提供支持。

**任务定义**：在此背景下，该评测任务旨在从互联网上真实存在的新闻数据中抽取特定类型事件的主体。即给定一段文本 T，和事件类型 S，从文本 T 中抽取事件类型为 S 的事件主体 E（如果事件主体为多个，输出事件主体 E 的集合）。形式化表示如下：

输入：一段文本 T 和事件类型 S

输出：事件主体 E（E 的集合）

示例：

*输入："公司 A 涉嫌违规交易，其下属子公司 B 和公司 C 遭到了调查"，"交易违规"*

*输出："公司 A"*

**任务难点：** 事件抽取一直以来是信息抽取中的难点问题，面向金融领域的事件主体抽取同样面临着诸多难点和挑战：1.语言表达的多样性，金融领域事件主体的实体类型一般为公司、机构或者人物，然而，数据来源广泛、表述风格不一的文本中，事件主体经常会以别称、简称或者代称的形式出现，语言表达的多样性对事件主体的识别提出了挑战。2.多事件表达的干扰性，一段文本会经常提到多个事件，一个实体也会同时参与到多个事件中。因此，如何从同时含有多个事件的文本中精确的找到特定类型事件的主体是一个挑战。3.事件主体的共现性，在金融领域的文本中会经常涉及多个相同类型事件的比较和统一表述，因此在同一段文本中同一类型的事件主体会有多个的情况，如何同时精准的识别出同一类型事件的多个主体是该任务的又一挑战。

**任务的组织及参赛情况：** 该评测任务采用 Biendata 在线评测平台，任务共分为 A 榜、B 榜两个阶段，A 榜测试阶段共持续 3 个月，A 榜阶段发布了训练集和测试集 A，每支队伍每天可以提交不超过 3 次的测试结果，在线测试平台可以实时更新队伍的最新排名。B 榜阶段共持续 4 天，每支队伍每天可以提交不超过 3 次的测试结果，在线测试平台可以实时更新队伍的最新排名，最终排名以 B 榜阶段测试成绩和提交报告说明为准。这次比赛共计吸引了 1169 名参赛者，487 支参赛队伍，包括：清华大学、北京大学、上海交通大学、复旦大学、浙江大学、哈尔滨工业大学、北京航空航天大学、北京邮电大学、哥伦比亚大学、曼彻斯特大学、香港理工大学、悉尼大学、芝加哥大学、香港城市大学等高校；中国科学院计算技术研究所、中国科学院软件技术研究所、中国科学院上海高等研究院、中国航空综合技术研究所、中国运载火箭技术研究院等科研机构；微众银行、平安科技、民生科技、腾讯、百度、谷歌创新工场、美团、字节跳动、京东、微软、IBM、国泰君安、网易等互联网和金融企业，具体参赛人员单位比例如图 7 所示。

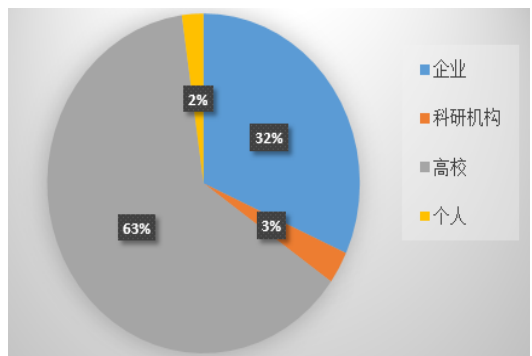

**图 7 参赛人员单位统计**

### 4.2 数据集及评估方法

**数据集来源：** 该评测数据集的文本内容来自于互联网上金融领域的相关新闻语料。数据爬取后首先进行数据清洗和数据过滤，主要是去掉无效的字符和内容。

**数据集构建方法：** 该评测数据由专业的标注人员从文本段中标注出事件类型和对应的事件主体（同一段文本中若出现多个事件类型和多个事件主体分别标注〈单一事件类型，多个事件主体〉的组合）。标注数据经过多个标注人员的标注和投票，最后进行抽样检验。

**数据统计：** 该评测数据集中共定义了"业绩下滑"、"提现困难"、"交易违规"等 21 个事件类型，如果事件不属于预定义的 21 个事件类型则标注为"其它"类型。训练集、A 榜阶段测试集和 B 榜阶段测试集中的样本分布如表 10 所示，各数据集中不同事件类型的样本分布如表 11 所示。在表 10 和表 11 中 Train 代表训练集、Eval 代表 A 榜测试集，Test 代表 B 榜测试集，Test_pure 代表 B 榜测试集中去除噪音数据的样本数量统计。

**表 10 不同数据集样本数量统计**

| 数据名称 | 样本数量 |
|---|---|
| Train(训练集) | 17815 |
| Eval(A 榜测试) | 3500 |
| Test（B 榜测试） | 135519 |
| Test_pure(B 榜测试) | 6988 |

**评估方法：** 本次任务采用精确率（Precision，P）、召回率（Recall，R）、F1 值（F1-measure，F1）来评估事件主体的识别效果。

事件主体精确率=识别事件主体与标注相同/识别事件主体总数量

事件主体召回率=识别事件主体与标注相同/标注事件主体总数量

事件主体 F1 值=(2*事件主体精确率*事件主体召回率)/(事件主体精确率+事件主体召回率)

### 4.3 典型模型与系统

本次评测中参赛队伍将面向金融领域的事件主体抽取任务建模成序列标注和阅读理解两种任务，最终排名靠前的队伍会在上述两种建模任务基础上利用模型集成，提升系统的鲁棒性和性能。下面分别介绍基于序列标注的方法和基于阅读理解的方法，然后再简要介绍模型集成的方法。

表 11　各数据集中不同事件类型的样本数量统计

| 事件类型 | Train | Eval | Test | Test_pure |
|---|---|---|---|---|
| 业绩下滑 | 686 | 148 | 0 | 0 |
| 提现困难 | 56 | 10 | 6520 | 570 |
| 失联跑路 | 43 | 9 | 4470 | 342 |
| 资产负面 | 116 | 25 | 1434 | 203 |
| 交易违规 | 1732 | 330 | 0 | 0 |
| 产品违规 | 62 | 12 | 1098 | 107 |
| 歇业停业 | 34 | 7 | 7066 | 331 |
| 不能履职 | 1326 | 247 | 0 | 0 |
| 涉嫌传销 | 500 | 104 | 8254 | 2033 |
| 公司股market异常 | 2 | 0 | 6 | 6 |
| 其他 | 3141 | 598 | 0 | 0 |
| 涉嫌违法 | 616 | 134 | 32775 | 412 |
| 财务造假 | 592 | 113 | 0 | 0 |
| 涉嫌非法集资 | 1644 | 333 | 9665 | 1608 |
| 资金账户风险 | 312 | 60 | 0 | 0 |
| 重组失败 | 1045 | 238 | 0 | 0 |
| 投诉维权 | 96 | 18 | 38854 | 159 |
| 高管负面 | 215 | 39 | 1280 | 306 |
| 信批failed违规 | 2513 | 497 | 0 | 0 |
| 评级调整 | 861 | 155 | 0 | 0 |
| 实控人股东变更 | 1827 | 345 | 0 | 0 |
| 涉嫌欺诈 | 396 | 81 | 23754 | 911 |

**基于序列标注的方法**：在该类方法中，参赛选手将面向金融领域的事件主体抽取任务建模成序列标注任务。输入为待抽取文本和事件类型，该类方法针对输入文本中的每个字进行标注，判断其是否为符合条件的事件主体，一般采用 BIO 的标注模式。在进行序列标注时，有两种建模事件类型的方法，一种是将 22（含其他类）个事件类别进行向量化表示，然后将其拼接到字向量后作为序列标注的输入；另一种是将事件类型的描述文本直接进行文本拼接，然后直接输入到序列标注模型进行训练。在参赛的队伍中，采用最多的序列标注模型为 BiLSTM+CRF[24]和 Bert[12]模型。

**基于阅读理解的方法**：在该类方法中，参赛选手将面向金融领域的事件主体抽取任务建模成

阅读理解任务。将输入文本看作阅读理解任务中的文档，将事件类型看作阅读理解任务中的问题。典型的方法是基于 BERT 的阅读理解模型，在该类方法中首先在事件类型（问题）前面添加 special classification token[CLS]标记，然后问题和段落连接处使用 special  tokens[SEP]分开。序列通过 token Embedding、segment embedding 和 positional embedding 输入到 BERT。最后，通过全连接层和 softmax 函数将 BERT 的最终隐藏状态转换为答案跨度的概率。

**基于模型集成的方法**：基于阅读理解的方法和基于序列标注的方法各有优势，基于序列标注的方法能充分捕捉多个事件主体之间的依存关系，基于阅读理解的方法能更好的挖掘文本和事件类型的语义关系。在单一模型中，不同的模型参数或模型设置也会得到不同的性能，为了充分利用多模型的优势，增加系统的鲁棒性，很多参赛队伍都采用了模型集成的方法，主要分为两类方法，一类是基于阅读理解的模型和基于序列标注模型的融合，一类是不同参数的单一模型融合。在此次比赛中，模型融合的方法多采用结果融合的方法，即取多个模型结果的并集。

### 4.4 评测性能及结果分析

此次评测的 A 榜和 B 榜 TOP 20 结果如表 12 和表 13 所示。详细比赛结果参见：
https://www.biendata.com/competition/ccks_2019_4/

表 12 A 榜 TOP 20 队伍成绩

| 排名 | 队伍名 | F1 值 |
|---|---|---|
| 1 | 我是谁我从哪里来我要去哪里 | 0.93917 |
| 2 | DOTA | 0.93819 |
| 3 | L | 0.93464 |
| 4 | 哪吒 | 0.93335 |
| 5 | 西南交大一枝花儿 | 0.93027 |
| 6 | BasinSpace | 0.92925 |
| 7 | 糯米糍 | 0.92864 |
| 8 | Axxx | 0.92833 |
| 9 | xiayang321 | 0.92809 |
| 10 | 炸鱼薯条 | 0.92648 |
| 11 | MSKJ 站上 KG 星球之巅 | 0.9262 |
| 12 | c_c | 0.92599 |
| 13 | 落霞与孤鹜齐飞 | 0.92597 |
| 14 | 秋水共长天一色 | 0.92587 |
| 15 | GDUFSER | 0.92584 |
| 16 | 蓝光剑侠天晴晚 | 0.9258 |
| 17 | bdy | 0.92514 |
| 18 | zc1 | 0.92479 |
| 19 | hihishenxian | 0.92445 |
| 20 | Dragon | 0.92426 |

通过对上述 A 榜和 B 榜的分析，我们发现：
（1）在单一模型中，基于阅读理解的方法性能普遍比基于序列标注方法的性能高，造成这种现象的原因可能是该次比赛中一个事件对应多个主体

的样本少于一个事件对应一个主体的样本。

（2）基于模型集成的方法更鲁棒，基于模型集成的方法在 B 榜上性能的降幅小于基于单一模型的方法的性能降幅。

表 13 B 榜 TOP 20 队伍成绩

| 排名 | 队伍名 | F1 值 |
|---|---|---|
| 1 | 糯米糍 | 0.83902 |
| 2 | GDUFSER | 0.83784 |
| 3 | L | 0.83784 |
| 4 | hihishenxian | 0.83572 |
| 5 | DOTA | 0.83503 |
| 6 | JJ | 0.83445 |
| 7 | konroy | 0.83248 |
| 8 | NLPxiaoxu | 0.8319 |
| 9 | justin | 0.83146 |
| 10 | 国士无双 | 0.83129 |
| 11 | Brad | 0.83129 |
| 12 | 我不会武功 | 0.83083 |
| 13 | 哪吒 | 0.83062 |
| 14 | 牛杂 | 0.82993 |
| 15 | 炸鸡薯条 | 0.82991 |
| 16 | 我是谁我从哪里来我要去哪里 | 0.82902 |
| 17 | MSKJ 站上 KG 星球之巅 | 0.82816 |
| 18 | zcl | 0.82809 |
| 19 | yaohua | 0.82769 |
| 20 | 便利贴 | 0.8264 |

### 4.5 任务总结

事件知识在金融领域有着不可或缺的地位，面向金融领域的事件主体发现是一个富有挑战的任务，该任务可以建模成序列标注任务，也可以建模成阅读理解任务，大规模预训练语言模型的引入会使相关方法进一步提升性能。在该任务建模过程中要重点考虑事件类型的编码方式以及事件类型和表述文本之间的语义关系。除了事件主体以外，事件发生的时间、涉及的金额和产品对于金融风控和智能投顾也有借鉴意义，未来可以尝试将该任务进一步扩展，抽取出事件主体以外的更多事件知识。

## 5 任务五：公众公司公告信息抽取

### 5.1 任务介绍

随着金融科技的发展和全球资本市场的不断扩大，在金融领域，每一天都有海量的数据产生，而与之形成强烈对比的是有限的人力以及人脑所能处理信息的极限能力。因此，依靠传统的人工方式已经无法应对投研分析、风险控制、金融监管和事件关联等需求，而亟需引入新的技术来提高信息处理效率，包括大数据分析、自然语言处理、知识图谱等技术，都已经开始被积极用于金

融分析和金融监管领域。在监管方面，每一家公众公司都具有相关信息披露义务，由此也产生了大量的公告阅读和信息抽取需求。据不完全统计，以沪深股市为例，2017 年共披露公告 44 万余篇，2018 年共 27 万余篇，并且随着上市公司数量的增加这一数字也在逐年增加。每年 3 月底、4 月底、8 月底、10 月底为定期报告披露高峰期，最多的一天所发布公告达 10297 篇。

本次评测的主要目标是针对公告文件（均以PDF 方式发布）中的信息抽取。作为知识图谱构建的基础，结构化数据是必不可少的。由此，如何通过自动化的技术来从各类公告中抽取信息，将非结构化数据转化为结构化数据是知识图谱领域所面临的一大挑战。

### 任务定义

本评测包含有两个子任务：

（1）表格中的信息点提取：本任务给定某公众公司的年报 PDF 文件，参赛者需从文件的结构化财务报表（包括合并资产负债表、母公司资产负债表、合并利润表、合并现金流量表和母公司现金流量表等）中提取相关的信息点，并输出该表格对应的结构化数据。

（2）文本段落中的信息点提取：本任务要求参赛者从公告中提取相关的重要信息点，采用的数据皆为"人事变动"类型的公告，其中可能包含离职高管、继任者等人物的相关信息，参赛者需要从中提取出相关的信息点，并输出对应的结构化数据。

两个任务的类型相似，其统一形式化定义可以被表示为：

输入：

1. 公众公司的年报文档集合/公众公司的人事变动公告文档集合：

$T = \{t_1, t_2, t_3, \cdots, t_n\}$

2. 预先定义的信息点类别：

$K = \{k_1, k_2, k_3, \cdots, k_n\}$

输出：

对于每个文档，输出其包含的所有信息点类别对应的信息。

$Output = \{(t_1, kv_1), (t_2, kv_2), (t_3, kv_3), \dots, (t_{n1}, kv_{n1})\}$，其中 $kv_i = \{(k_1, v_1), (k_2, v_2), (k_3, v_3), \dots, (k_{n2}, v_{n2})\}$

### 任务难点

本任务为公众公司公告信息提取，两个任务均需要先从结构化的数据中定位有效信息段落，再将关键信息进行提取并结构化。其主要难点在于：

PDF 文件的信息定位和提取。PDF 类型的文件是按字符和位置存储的，缺少其内容的结构化信息，段落和表格之间没有明显、清晰的边界。该特性增大了对其信息自动抽取的难度，同时也提高了对性能的要求。因此，如何准确地保留 PDF 文档中的显式的结构化信息，如段落、表格等，并减少噪声是解决该问题的重点和难点。

表格信息提取。与传统的文档信息抽取不同，表格的前后文本内容中也包含了相关的信息，要深入挖掘表格的信息，需要结合其上下文的内容，进行综合分析。

信息的不确定性。并非每个公告都会包含所有有关数据，部分人事变动仅有离职者而没有继任者，模型在尽可能提取相关信息的同时，还需要考虑特定键值是否缺失的情况。

**任务组织**

本次测评由东南大学认知智能研究所组织，时间安排如下：

- 评测任务发布：4 月 1 日
- 报名时间：4 月 1 日 — 4 月 20 日
- 训练及验证数据发布：4 月 20 日
- 评测文件提交：7 月 20 日
- 评测时间：7 月 25 日
- 评测论文提交：8 月 15 日

本次评测依托 bien-data 平台展开，评测网站链接为：[https://biendata.com/competition/ccks_2019_5/](https://biendata.com/competition/ccks_2019_5/)。本次测评训练数据为公开发布的公众公司定期报告文件。组织者将提供训练数据集和验证集（包括公告 PDF 原文和对应的结构化数据），供参赛选手训练算法模型和参与验证排名。评测数据为训练数据集和验证集同一类型公告，为防止作弊和人工介入，评测数据将包含非公开发布的年报和公告 PDF 文件（人为制造），评测数据不提前进行发布。参赛者须提供可调用的 API，组织者将基于评测数据在指定评测时间统一调用 API 来给出最终分数。

评测过程中，组织者对参赛者的 API 调用处理时间和结果返回时间设置上限，若一条评测数据未能在限定时间内返回结果，该评测数据及对应的信息点将按抽取失败计入评价指标计算中。

**参赛情况**

本次测评共有 98 支队伍，共 318 名参赛者参加。其中企业参赛队、研究机构参赛队分别有 xx、xx 支。经过测评，提交有效结果的队伍有 xx 支，其中有 6 支队伍提交了测评论文。

**5.2 数据集及评估方法**
**数据描述**

本次测评训练数据为公开发布的公众公司定期报告文件。组织者将提供训练数据集和验证集（包括公告 PDF 原文和对应的结构化数据），供参赛选手训练算法模型和参与验证排名。评测数据为训练数据集和验证集同一类型公告，为防止作弊和人工介入，评测数据将包含非公开发布的年报和公告 PDF 文件（人为制造），同时，评测数据不提前进行发布。参赛者须提供可调用的 API，组织者将基于评测数据在指定评测时间统一调用 API 来给出最终分数。

**评价指标**

两个子任务的评价均采用传统的正确率（Precision）、召回率（Recall）和 F1 值作为评价指标。将文本集合记为 $T=\{t_1, t_2, t_3, \cdots, t_n\}$，对于文本 $t_i$，记该文本中的信息点个数共 $G_i$，记模型提取出的信息点个数共 $A_i$，其中的正确信息点个数记为 $N_i$，相关计算公式如下：

$$\text{Precision} = \frac{1}{|T|}\sum_{i=1}^{|T|} P_i, \ P_i = \frac{|N_i|}{|A_i|}$$

$$\text{Recall} = \frac{1}{|T|}\sum_{i=1}^{|T|} R_i, \ R_i = \frac{|N_i|}{|G_i|}$$

$$\text{F1} = \frac{1}{|T|}\sum_{i=1}^{|T|} \frac{2*P_i*R_i}{R_i+P_i}$$

**5.3 典型模型与系统**

本次评测的参赛者共提交 7 篇论文。由于 PDF 数据的特殊性，大部分队伍将解决该问题的方案分成了两个模块，即，先是从 PDF 中自动抽取相关的信息，再对这些信息做进一步的分析与挖掘。通过对这两个模块分别进行训练和优化，模型得以输出更为准确的结构化数据。分析这些论文后，我们将其采用的典型方法与模型总结如下：

**5.3.1 PDF 信息自动抽取模型**

解决 PDF 信息自动抽取主要有两种思路，概述如下。第一种是先将 PDF 文件转换为图片，再找到所需区域，并对其进行限定和提取。与队伍利用 BFS 算法对表格内内容进行定位和提取[25]，并准确定位到表格提取的难点-跨页表格中信息的准确识别，通过构建页眉页脚识别模型，有效地解决了该任务难点；也有队伍采用 OpenCV 接口和 Faster R-CNN 算法，分别检测有线条和无线条的表格，同样获得了不错的效果[28]。第二种方法基于工具将 PDF 文件转换为设定的格式以获取结构化信息。有队伍使用 Acrobat DC SDK，将 PDF 文档转换为 XML 文件，成功识别了大部分的文字和表格[27]，但该方法会将部分表格错误地识别为图片，降低了模型的效果。

### 5.3.2 混合模型

任务二可被作为序列标注问题来进行建模，按照这个思路，需要先解决命名实体识别问题。大部分队伍都采用了 BiLSTM-CRF[24] 的神经网络模型完成命名实体识别。首先基于词向量进行字/词嵌入操作，再通过 BiLSTM 层提取文本特征，最后通过 CRF 层给每个单位打上标签。混合模型的效果相较于单个模型有显著的提高。

### 5.3.3 预训练语言模型

近来，基于大量无监督文本的深度神经网络预训练模型大幅地提升了各个 NLP 任务的模型效果。在本评测任务中，参赛者通过使用包括 BERT[12] 在内的多种预训练模型，显著地提升了模型的效果。其中，在解决任务二时，有队伍通过使用 BERT，基于上市公司公告进行了预训练，构建了针对证券公告领域的语料模型 caBERT[26]，并针对"人事变动"类型公告进行了 fine-tune。通过使用 BERT-CRF 和 caBERT，其模型的 F1 值比使用最常用的 BiLSTM-CRF 模型提高了 13.14%。最终，该模型达到 95.78% 的 F1 值，在该单向任务中排名第一。

**表 14 子任务 1（年报抽取）评测结果**

| 排名 | 参赛队名 | 单位 | F1 平均值 |
|---|---|---|---|
| 1 | 美能华 | 苏州美能华智能科技有限公司 | 0.99750 |
| 2 | DG | 达观数据 | 0.97810 |
| 3 | 洞见时代 | 洞见时代（北京）信息技术有限公司 | 0.92477 |
| 4 | guangluwutu | 中国地质大学（武汉）计算机学院 | 0.89055 |
| 5 | DataHammer | 北京理工大学 | 0.88711 |
| 6 | ChenXiuling | 平安科技 | 0.82787 |
| 7 | louis_xu | 新氢数据 | 0.82423 |
| 8 | NiHao 文本分析 | 大连理工大学自然语言处理实验室 | 0.74856 |

**表 15 子任务 2（人事公告抽取）评测结果**

| 排名 | 参赛队名 | 单位 | F1 平均值 |
|---|---|---|---|
| 1 | SZSI | 深圳证券信息有限公司 | 0.95779 |
| 2 | 美能华 | 苏州美能华智能科技有限公司 | 0.94118 |
| 3 | DG | 达观数据 | 0.93994 |
| 4 | NiHao 文本分析 | 大连理工大学自然语言处理实验室 | 0.88800 |
| 5 | DataHammer | 北京理工大学 | 0.87272 |
| 6 | louis_xu | 新氢数据 | 0.86685 |
| 7 | guangluwutu | 中国地质大学（武汉）计算机学院 | 0.83445 |
| 8 | 洞见时代 | 洞见时代（北京）信息技术有限公司 | 0.74214 |

### 5.4 评测性能及结果分析

最终评测结果的前八名如表 14、15 所示。通过对参赛队所提交的测试借口调用、结果验证、文档及评测论文的对比审核，评测结果分析如下。

1）大部分参赛队伍的接口都出现调用超时的情况。经过逐一和参赛队伍校对，包括获取源代码进行本地测试，发现其中模型计算效率较差是造成很多测试样例信息点抽取失败的原因。如何快速的进行 PDF 文件解析以及抽取信息点，保证抽取速度，依然是较大的挑战。测试成绩较好的队伍，都不同程度对基于神经网络的模型进行了压缩和优化。

2）两项任务的总冠军和任务二的单项冠军，都针对不同领域在句法和文法上的差异基于 Bert 预训练生成领域词向量，将序列-触发词作为共同输入来触发对应元数据信息点的抽取，从而提高 PDF 文件信息点抽取的召回率和准确率。这一点具有一定的理论和实际应用价值。值得相关任务借鉴采纳。

3）基于 OpenCV 和 Faster R-CNN 的表格信息抽取方法，能够有效提高扫描插件的线条检测精度，以及无线条表格的单元格识别准确度。在此基础上，结合"表格定位+表格结构识别+信息点抽取"的三步策略识别出来表格区域中的信息点，是提高表格信息抽取模型效果的有效途径。

### 5.5 任务总结

PDF 格式的文档因其优秀的展示效果，在多种场景下被广泛地采用，如：科学论文、官方文件、演示文稿等，显然地，该类文档中含有海量的高质量信息。然而因其侧重不同，该类文档中的数据并未被结构化储存，在过去，人们若要提取出 PDF 文档中的信息，一般只能通过人工方式手动进行提取。这种方法效率低下，且提取后的信息还需要重新编辑、排版。因此，如何准确高效地自动提取 PDF 文件中的信息，并将其结构化储存，是知识图谱领域中的一个重要课题。

本评测希望通过将 PDF 信息抽取与下游任务相结合，为后续的相关任务打下基础。同时，在本次任务中，我们所采用的数据来自上市企业披露的相关文件。批量地提取并分析这类文件的信息，可以显著地提高相关领域从业人员的工作效率和工作质量。

在本任务中，参赛者们采用了不同思路的方案解决了其中的难点，同时也使用了不同的模型建构思路提升了系统的效果。本次比赛促进了对 PDF 中元数据抽取的相关领域研究，拓宽了学术界的视野。我们在后面的工作中，会继续分析垂直行业的具体需求，提出更有实用意义和研究价值的评测人物，以促进知识图谱领域技术的进步。

# 6 任务六：中文知识图谱问答

## 6.1 任务介绍

### 任务定义

本评测任务是，对于给定的一句中文问题，问答系统从给定知识库中选择若干实体或属性值作为该问题的答案。问题均为客观事实型，不包含主观因素。理解并回答问题的过程中可能需要进行实体识别、关系抽取等子任务。这些子任务的训练可以使用额外的资源，但是最终的答案必须来自给定的知识库，并且使用的额外资源必须明确说明来源。为了方便参赛队伍训练模型，我们专门为本次测评构建了专门的中文自然语言问答数据。参赛队伍可以自由选择是否使用提供的训练数据。

为了保证比赛结果的公平、公开、公正，参赛队伍的成绩将实时公布并排名，比赛截止后，参赛队伍必须提供最好结果的实现代码以及相关数据，供赛方进行复现。否则，将被取消最终资格。

### 任务难点

本任务为开放领域的自然语言问答，宏观而言，这是一个从理解自然语言，再利用获取到的信息匹配知识库中数据的过程。其主要难点有：

语义消歧，在自然语言中，词语的意思往往是多种多样的，要结合具体的上下文来确定词语的含义。并且有时就算结合了上下文信息，歧义也依旧存在。在中文中尤其如此。

句法的模糊性。自然语言的文法通常是模棱两可的，针对一个句子通常可能会解析出多种依赖结构，而我们必须要仰赖语意及前后文的信息才能在其中选择一种最为适合的句法结构。

问题逻辑的复杂性。当自然问题涉及复杂的逻辑推理和联系时，需要知识库中距离更远的信息。

## 任务组织与参赛情况

本次测评由北京大学计算机科学技术研究所与恒生电子股份有限公司共同组织，共有 165 个队伍参加。其中企业参赛队、研究机构参赛队分别有 33、74 至。经过测评，提交有效结果的队伍有 28 支，其中有 4 支队伍提交了测评论文。

## 6.2 数据集及评估方法

### 数据来源

本次测评任务使用北京大学开发的 PKUBASE 作为指定的知识图谱。其被用于数据集的生成，并且选手最终给出的问题答案必须来自于该知识库。PKUBASE 采用当前应用最广泛的 RDF 结构，用户可以使用结构化的 SPARQL 语言对其进行查询。因此，在训练数据集中，也包括了人工给出的问题对应的 SPARQL 查询。选手可以下载数据集并使用相应的知识库管理系统对数据进行存储和查询，我们也提供了在线的 PKUBASE 查询终端，供选手通过浏览器或者 API 进行在线访问。详情可见 [http://pku-base.gstore-pku.com/](http://pku-base.gstore-pku.com/)

### 数据生成

本次测评任务构建了专门的中文自然语言问答数据集。数据集问题与答案生成过程大致分为如下几步：第一步，人工确定自然语言问题模板。我们认为，利用 PKUBASE 这样 RDF 格式的开放知识库生成问题及其答案，首先要将自然语言问题与 SPARQL 查询关联起来。因此，我们首先确定了几种典型的 SPARQL 查询图的模板，利用这些查询图的模板，补全对应的实体与谓词信息，再进一步生成自然语言问题。这一步是由在自然语言问答与开放知识库领域具有专业知识的人来完成。我们用到的问题模板包括单跳型、双跳型、三跳型以及星型。我们将其中的"单跳问题"模板称为简单问题模板，而其余的均为复杂问题模板。需要注意的是，虽然我们

利用 SPARQL 查询作为生成问题的中间状态，我们并不强制要求参赛队伍再开发系统时将问题转化成 SPARQL 查询，再得到答案。

第二步，人工利用问题模板生成中文自然语言问题。这个过程就是将模板转化为具体的 SPARQL 查询，再转化成自然语言问题。我们邀请了共计 22 位来自北京大学的工作人员，他们均为中文母语者，具有较高的中文语言水平，并且对 RDF 和 SPARQL 查询有所了解与研究，因此可以保证得到的问题的质量。我们要求每位工作人员需要编写 100 道自然语言问题，并且给出对应的答案。其中，简单问题与复杂问题的比例为 1:1，五种不同模板的比例为 5:2:1:1:1。

在这一过程中，我们建议工作人员首先根据模板，从 PKUBASE 中任意搜索实体，根据实体周围的谓词补全模板，同时设置查询变量与查询变量，并且只能设置一个查询变量。以简单问题为例，工作人首先查询所有与"〈湖上草〉"这一实体相连的实体及其谓词，并发现存在"〈柳如是_(明末"秦淮八艳"之一)〉 〈主要作品〉 〈湖上草〉"这一三元组，满足简单问题的模板。这时，三元组中的主语"〈柳如是_(明末"秦淮八艳"之一)〉"可以被视为查询变量，进而得到如下的 SPARQL 查询：

```
select ?x where {
  ?x 〈主要作品〉 〈湖上草〉.
}
```

然后，工作人员将 SPARQL 查询改写为自然语言问题，例如上述 SPARQL 可以改写成"《湖上草》是谁的诗歌？"，也可以改写成"谁写了《湖上草》？"。我们要求工作人员在改写过程中，对同样的模板尽量使用不同的句式，并且加入指代、歧义等自然语言现象，以保证数据集的质量和复杂程度。最后，问题的答案直接使用对应的 SPARQL 查询得到的结果来充当。

### 数据清洗与分割

为了保证数据质量，我们对生成的 3847 组中文自然语言问答数据进行了数据整理与清洗。具体方法是，首先对所有 SPARQL 查询进行验证，将每一条查询提交给我们提供的在线查询终端，测试其是否能正常查询到答案，以保证其格式正确，并且能够从 PKUBASE 中得出准确的答案。然后，人工对 3847 条自然语言问题进行审核，保证其符合中文语言规范和语言习惯。

经过数据清洗后，我们总计得到了 3830 组符合要求的中文问答数据，并且其简单问题与复杂问题的比例依旧保持在 1:1 左右，五种模板的比例保

持在 5:2:1:1:1 左右。随后，我们采用随机抽取的方法将 2083 组数据划分为训练集、验证集与测试集。其中，训练集包含 2298 组，验证集与测试集各 766 组，并且保证问题类别与问题模板的比例与前文提及的一致。

### 评价方法

本任务的评价指标包括宏观准确率（Macro Precision），宏观召回率（Macro Recall），Averaged F1 值。最终排名以 Averaged F1 值为基准。设 $Q$ 为问题集合，$A_i$ 为选手对第 $i$ 个问题给出的答案集合，$G_i$ 为第 $i$ 个问题的标准答案集合，相关计算公式如下：

$$\text{Macro Precision} = \frac{1}{|Q|}\sum_{i=1}^{|Q|} P_i,\ \ P_i = \frac{A_i \cap G_i}{|A_i|}$$

$$\text{Macro Recall} = \frac{1}{|Q|}\sum_{i=1}^{|Q|} R_i,\ \ R_i = \frac{A_i \cap G_i}{|G_i|}$$

$$\text{Average F1} = \frac{1}{|Q|}\sum_{i=1}^{|Q|} \frac{2P_iR_i}{R_i+P_i}$$

### 6.3 典型模型与系统

大多数取得较好表现的队伍将生成答案的系统分成多个功能模块，分别训练或优化每个模块并构建 pipeline 以完成最终的查询语句生成或答案选取。诸多功能模块中，最常被使用的包括核心实体识别、关系识别、问题类型识别、路径匹配与排序等，针对这些核心模块的优化方法成为了部分队伍能获得优异表现的主要原因。

本次比赛，表现较好的几个系统可以大致分为两类，其一是通过实体识别、关系抽取获得答案，其二是通过实体识别、路径匹配选择答案，接下来将分别概述这二者。

#### 6.3.1 基于实体与关系识别的模型

模型的流程图如图 8 所示。此类模型的第一步是通过 NER 等方法获得问句中实体的 mention，再通过主办方提供的数据库中 mention2ent 及其他信息链接 mention 到数据库中对应实体列表，通过多种设计的打分函数最终选取一个或多个与问句最相关的实体作为核心实体，用于之后生成关系、查询语句等。在 NER 模型方面，有的队伍使用了在线训练好的模型[29]，也有队伍使用 BERT-BiLSTM-CRF 并微调以提高其在此任务中的表现[30]；在实体打分函数方面，简单到实体长度，复杂到实体与问句语义相关性（须通过模型训练）均被不同队伍使用。

获得中心实体后，此类系统先得到该实体在数据库中的邻居子图，并抽取出邻居子图中所有的关系作为候选。对于这些关系，系统同样使用设计好的打分函数来评估每一者与问题的契合性，这些打分方式包括关系本身与问题语义相似性及对应询问路径中相关结点与问句的相似性等，使用到的模型

也包括 BERT、BiLSTM[29]等。

同时，此类模型也大多采用一个问题类型识别模块以提高表现。数据集中的问题可以分为单跳问题与多跳问题，多跳问题在查询图结构上也有多种形式；通过训练，问题类型识别模型能够将问句分类到相应的查询图结构模板上。这些有了这些模板信息，便可以根据之前获得的关系取打分高的一个（如果问题类型为单跳）或多个（如果类型为多跳）作为结果，将其与中心实体对应的查询路径填入最终查询数据库得到答案。

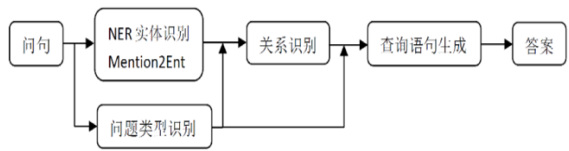

图 8 基于实体与关系识别的模型总体流程

### 6.3.2 基于路径匹配的模型

模型的流程图如图 9 所示。与上一类模型类似，此类模型同样具有识别中心实体、问题类型的模块。不同的是，本方法并不直接获取关系，而是直接获取所有能匹配上问题类型模板的、在中心实体周边的查询路径；之后，通过各类基于问句与查询路径特征来给后者打分，并选取最优查询路径的模型，最终答案便可以求得。

一般而言，即便利用各类启发性剪枝，通过上述流程产生的候选查询路径是众多的，尤其当问题为多跳、查询路径较长时，因此，优化查询路径的打分方式对于此类系统是至关重要的。在本次比赛中，表现较为突出的模型包括 BERT 语义匹配模型[30]、Pairwise LambdaRank[31]等。

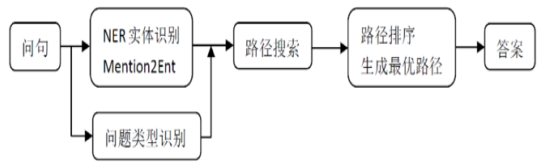

图 9 基于路径匹配的模型

除此之外，完全基于规则的传统系统也出现在本次比赛中，部分队伍亦使用 Ensemble 融合多个系统结果以试图达到单个系统无法企及的性能，各类不同的系统各有优劣，启发我们进一步探索。

## 6.4 评测性能及结果分析

### 6.4.1、最终结果

经过数月代码提交与测评验证，jch1 队最终名列第一，下表 16 为本次比赛最终排列前 10 名的队伍。

表 16 评测结果

| 排名 | 队伍名 | F1 得分 |
|---|---|---|
| 1 | jch1 | 0.73545 |
| 2 | hlt217 | 0.73075 |
| 3 | 网易互娱 AIlab-陈垚鑫 | 0.72514 |
| 4 | baseline | 0.70448 |
| 5 | DUTIR | 0.67683 |
| 6 | 到此一游 | 0.63510 |
| 7 | qbuer | 0.60566 |
| 8 | Duoduo 小分队 | 0.54227 |
| 9 | 单身公寓队 | 0.50658 |
| 10 | 我是一条鱼 | 0.42345 |

表 17 错误分类

| 错误类别 | 可能错误原因 | 示例 |
|---|---|---|
| 实体识别错误 | 某些 mention 可能同时指代不同的实体，需要利用句子上下文信息进行消歧 | 《神话》是哪位导演的影片？（《神话》可能是电视剧也可能是电影，需要靠上下文信息） |
| 谓词匹配不准确 | 句子中出现的文本与知识库中的谓词有时候差距很大；同一个谓词 mention 也可能对应许多个谓词，需要利用上下文信息综合确定。 | 《国王的演讲》有多少钟？（知识库内谓词为"片长"） |
| 查询路径形式错误 | 当自然语言问题涉及复杂逻辑推断，例如三跳问题，系统表现出额外的信息。大多数系统没有考虑长查询路径的信息。 | 尼采来自同一国家的哲学家都有那些？（问题共有三跳） |
| 非问句形式 | 有时祈使句也能当作自然语言问题，需要特别处理。 | 告诉我刘翔的身高。 |
| 不适应开放领域 | 对于问题相较少的领域，模型识别、理解的准确率降低。 | 如何计算融券费用？（融券费用实体不常涉及） |

### 6.4.2 结果分析

分析模型给出回答的错误原因往往是理解当前模型弱点的直接方式。下表 17 总结出了我们发现的几种主要回答错误类别，同时给出了可能的错误原因及典型致错问题示例。

可见，问句本身的复杂性仍然是导致错误的主要原因，自然语言的不确定性使得对实体、关系、类别的识别在很多问题上存在困难。对此，各参赛队使用多种模型提取上下文语义信息，结合多种基于规则的匹配方法综合判断，在处理问句复杂性的方面做出了值得借鉴的改进。

相比往年比赛结果，本次比赛在得分上有着较大程度的提高，这得益于语言模型 BERT[12]的提出及其预训练模型在诸多下游任务中的应用，相比起 LSTM、TextCNN 等上一代模型，BERT 尤其在实体、关系、问句类型识别上能达到更高的准确率。可见，积极应用自然语言处理领域的新兴模型、技术是有

利于进一步提高问答系统性能的。另一方面，在本次比赛中得分较高的队伍均采用模块化系统的思路，端到端的直接生成查询方法并未受到重视，这有可能是因为当前数据集中句量还不够庞大，并不足以提供足够的隐含信息让模型来学习和理解；对于KBQA任务，虽然模块化系统是直接的思路，也降低了训练的数据要求，但诸多在其他领域的研究发现端到端的模型利于计算机发现更多潜在的、人为难以识别的模式，反而有益于提高模型的性能。随着CCKS中文问答数据集的不断丰富，将端到端模型尝试应用与中文问答任务是值得期待的。

最后，大多数参赛队伍均将查询路径限定在2跳内，一些队伍还将查询路径模板进一步缩减，这一方法对于给定数据集可能是有效的：减小路径模板数有助于相应模型训练出更高的识别准确率，同时降低了计算开销；如果复杂问题占比较少，少数模板就可以覆盖大多数问题，这一做法可以用较少损失换取在简单问题上较大的性能提升，进而取得更高得分。然而，为了得到面向开放领域的通用知识问答系统，对复杂问题占比较少的假设并不一定成立，仅考虑2跳以内的问句也许会严重影响模型在实际应用中的性能。因此，增强模型面对更加复杂多样问题的健壮性是进一步提高模型的必要条件。

### 6.5 任务总结

针对开放领域的知识图谱问答，本次比赛的参赛者们提出了各类模型架构；面对此任务的难点，各队提出了不同解决方案，相比于往年评测结果切实地提高了问答系统地整体表现。本次比赛地拓展了学界对中文问答的认知，促进了中文知识图谱问答系统的继续发展。

继去年赛方首次提出中文知识图谱自然语言问答数据集后，此数据集进一步修正与扩充，增加了大量金融领域问题，这使得数据集复杂性增加，更加切合开放领域、复杂问题的实际情况，进而促进相关领域研究不断发展。

总之，本次测评取得一定的效果，我们会后续不断扩充中文知识图谱数据集的规模和多样性，为参赛者提供更有挑战更有意义的问题，也希望在今后的测评中各研究团队能继续突破，再创佳绩。

## 7 总结及展望

综合上述各个任务的综述，我们可以看到，目前已有非常丰富的知识图谱技术和工具，这些技术在很多方面都达到了应用的性能水平。具体地，我们可以看到：

1. **给定合适的监督语料和应用场景，当前知识图谱技术性能水平已经达到实用水平。** 我们可以看到，在本次评测中，排名前列的系统都取得了相当高的技术水平。且这些系统大部分是基于深度学习技术，同时具有易于构建和易于使用的特点；

2. **大规模预训练模型和模型集成是达到高性能的关键。** 在多个任务中，排名前列的系统基本都采用了大规模预训练模型（如BERT和Elmo）然后微调的模式。同时，多个系统通过集成学习进一步提升了系统的性能；

3. **领域知识资源可以帮助构建有效的模型，特别是在面向特定领域时。** 例如，在医疗命名实体识别中，医疗实体词典可以有效的降低对语料的需求并提升系统的性能。

同时，我们也看到，虽然知识图谱技术已经取得了长足的进展，但是知识图谱仍有许多值得研究的方向和问题：

1. **低资源的知识图谱技术需要进一步研究。** 虽然在有监督的情况下知识图谱技术性能水平优秀，但是在缺乏语料的情况下知识图谱距离实用仍有一定距离。考虑到语料标注的成本，在低资源情况下构建高性能知识图谱系统是一个非常值得研究的方向；

2. **在将知识图谱技术应用到实际任务中时，需要充分考虑领域、语料、任务等方面的差异。** 在本次评测中，当训练语料与测试数据集基于不同应用场景时，性能会有一个显著的下降。在实际应用中，训练语料场景和实际语料场景不一致往往是更为常见的场景。例如，在规范文本中训练，在口语文本中应用。这就需要我们研究可以考虑领域、语料和任务差异的鲁棒知识图谱技术，如迁移学习技术、无监督学习技术等；

3. **中文具有独有的特点，需要在设计知识图谱技术时充分考虑。** 中文具有没有分词标记、意合等特点，中文知识图谱技术往往比英文知识图谱技术更具挑战。

在未来工作中，CCKS将持续举办评测，并针对知识图谱技术的方向和水平持续改进。相信CCKS系列评测可以为提供构建知识图谱系统的技术参考，评估当前知识图谱技术水平，揭示未来的发展方向持续提供一个有价值的平台，并最终服务于知识图谱技术的介绍、交流、应用和推广。

# 参考文献